\pdfoutput=1

\documentclass[11pt]{article}

\usepackage[]{acl2023}

\usepackage{times}
\usepackage{latexsym}

\usepackage[T1]{fontenc}

\usepackage[utf8]{inputenc}

\usepackage{microtype}

\usepackage{inconsolata}

\usepackage{todonotes}
\usepackage[normalem]{ulem}
\usepackage{subcaption}
\usepackage{amsmath}
\usepackage[export]{adjustbox}
\usepackage{enumitem}
\usepackage{catchfile}
\usepackage{tikz-dependency}
\usepackage{booktabs}
\usepackage{hyperref}

\makeatletter
\newcommand*{\rom}[1]{\expandafter\@slowromancap\romannumeral #1@}
\makeatother

\newcommand{\hd}[0]{Hard-Debiasing}
\newcommand{\opus}[0]{OPUS-MT}   
   
\title{The Impact of Intrinsic Debiasing on Downstream Tasks:\\ A Case Study on Machine Translation}
\title{Applying Intrinsic Debiasing on Downstream Tasks:\\Challenges and Considerations for Machine Translation}

\author{Bar Iluz \textsuperscript{$\diamondsuit$}
, Yanai Elazar \textsuperscript{$\spadesuit$,$\clubsuit$}, Asaf Yehudai \textsuperscript{$\diamondsuit$}, Gabriel Stanovsky \textsuperscript{$\diamondsuit$}\\
\textsuperscript{$\diamondsuit$}Hebrew University of Jerusalem,  
\textsuperscript{$\spadesuit$}Allen Institute for AI,\\
\textsuperscript{$\clubsuit$}University of Washington\\
\\
\texttt{\{bar.iluz,gabriel.stanovsky,asaf.Yehudai\}@mail.huji.ac.il} \\
  \texttt{yanaiela@gmail.com}\\}

\begin{document}
\maketitle
\begin{abstract}

Most works on gender bias focus on intrinsic bias --- removing traces of information about a protected group from the model's internal representation. However, these works are often disconnected from the impact of such debiasing on downstream applications, which is the main motivation for debiasing in the first place.
In this work, we systematically test how methods for intrinsic debiasing affect neural machine translation models, by measuring the extrinsic bias of such systems under different design choices.
We highlight three challenges and mismatches between the debiasing techniques and their end-goal usage, including the choice of embeddings to debias, the mismatch between words and sub-word tokens debiasing, and the effect on different target languages. We find that these considerations have a significant impact on downstream performance and the success of debiasing.
\end{abstract}

\section{Introduction}
\label{sec:intro}
Natural language processing models were shown to over-rely and over-represent gender stereotypes.\footnote{Throughout this work we refer to morphological gender, and specifically to masculine and feminine pronouns as captured in earlier work. We note that future important work can extend our work beyond these pronouns to e.g., neo-pronouns~\citep{lauscher2022welcome}.} These can typically be found in their internal representation or predictions. For example, consider the following sentence:

\noindent\begin{dependency}
    \begin{deptext}
    (1) \& The \kern-2ex \& \textbf{doctor} \&\kern-2ex asked \& the \& \textbf{nurse} \& to \& help \& \textit{her} \\ in \& the \& procedure.\\
    \end{deptext}
    \depedge[edge height=0.5em]{9}{3}{coref}
    \label{ex1}
\end{dependency}
Inferring that \textit{her} refers to the nurse rather than the doctor may indicate that the model is biased.
A useful distinction of model's biases was proposed by \citep{goldfarb-tarrant-etal-2021-intrinsic,cao-etal-2022-intrinsic}:
\emph{Intrinsic bias} typically manifests in the geometry of the model's embeddings. 
For example, finding that stereotypically female occupations (e.g. ``nurse'', ``receptionist'') are grouped together in the embedding space, while stereotypically male occupations (e.g. ``doctor'', ``CEO'') are closer to each other \cite{gonen-goldberg-2019-lipstick}.
\emph{Extrinsic bias} on the other hand is measured in downstream tasks.
For instance, in machine translation (MT), which is the focus of this work, a biased model may translate Example (\ref{ex1}) to Spanish using a masculine inflection for the word ``doctor'', even though a human translator is likely to use a feminine inflection~\citep{stanovsky-etal-2019-evaluating}. 
Intrinsic and extrinsic bias do not necessarily correlate \citep{cao2022intrinsic,goldfarb2020intrinsic}, and biases might reoccur when applying debiased models on other tasks \citep{orgad2204gender}.


\begin{figure}[!t]
    \centering
    \includegraphics[width=0.6\columnwidth,center]{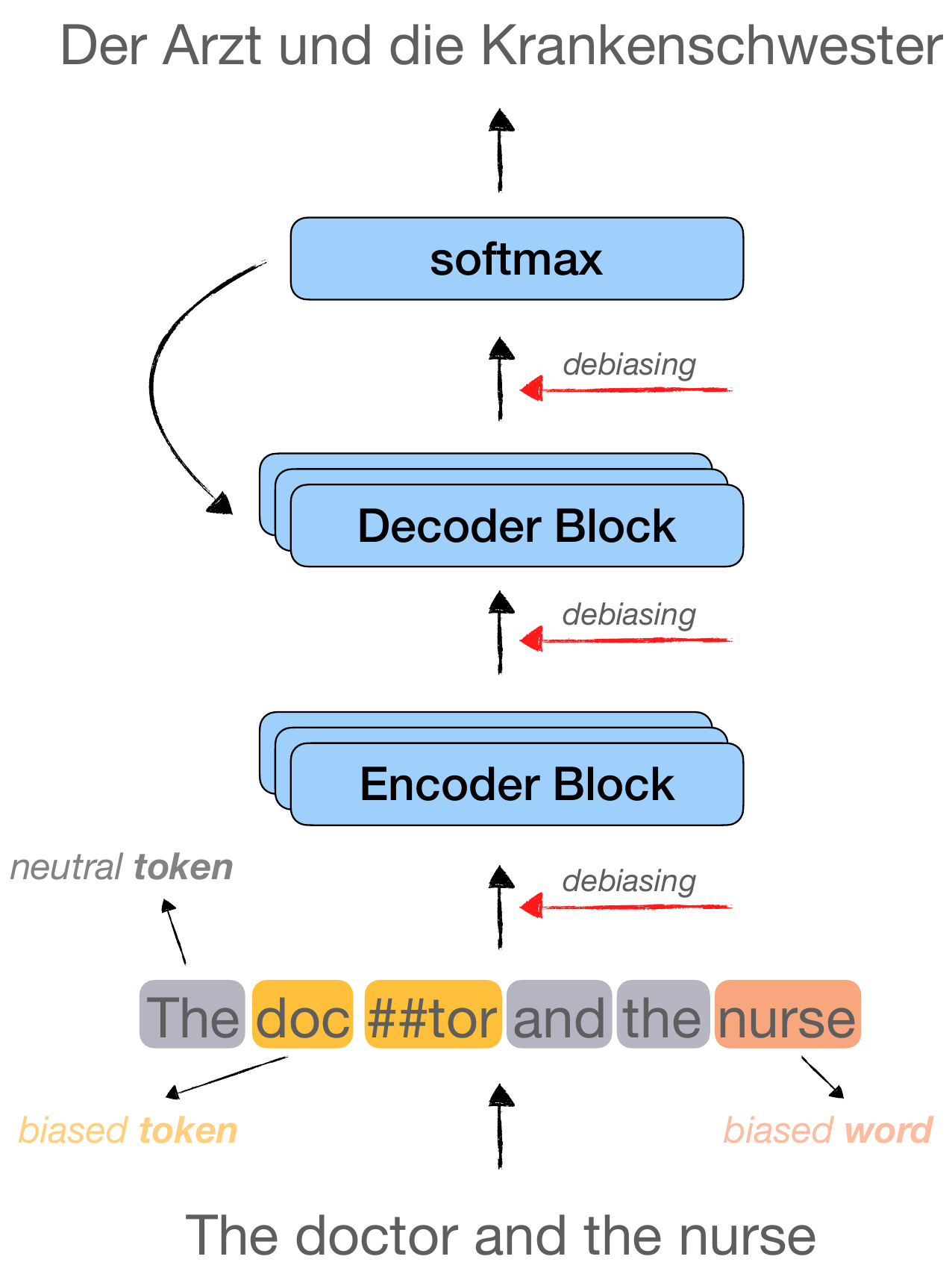}
    \caption{\label{fig:shwartz}A schematic view of a neural machine translation system, highlighting different possibilities for applying intrinsic debiasing techniques. We examine three considerations: (1) where to apply the debiasing; 
 (2) which tokens to apply the debiasing to (e.g. only gender-indicative words or the entire vocabulary); and (3) the effect of different target languages.}
    
\end{figure}

In this work, we identify a gap in the literature between intrinsic bias mitigation and its influence on downstream tasks.
Namely, while \emph{extrinsic bias} may affect human users in a variety of applications, debiasing techniques often focus only on \emph{intrinsic measures}, aiming to obfuscate gender from pretrained embeddings~\citep{bolukbasi2016man,elazar2018adversarial,ravfogel2020null}.
These approaches 
leave many unanswered questions when deploying them within a complex downstream model for specific tasks.

As shown in Figure~\ref{fig:shwartz}, we systemically explore three fundamental challenges when integrating intrinsic debiasing techniques within complex open-source neural MT architectures. We find that different design choices lead to a wide difference in extrinsic bias as well as task performance.

First, we explore different approaches to cope with discrepancies between different tokenization strategies. 
While intrinsic debiasing is largely performed over complete words from a fixed dictionary, modern MT requires mapping those onto sub-word elements determined via a data-dependant tokenizer. We find that debiasing only complete words outperforms a more naive debiasing of all sub-word tokens. 

Second, several word embedding tables could be debiased within an MT system.
Therefore, a preliminary architectural question is which of them to debias. We explore various combinations, finding the optimal configuration depends on the intrinsic debiasing technique.

Third, We explore the effects of debiasing a translation model over three target languages (Hebrew, German, and Russian). While all three encode morphological noun gender, they differ in script, typology, and morphology. We find that an important factor for debiasing efficiency is the number of words represented as single tokens, a property determined both by the language's morphological properties as well as its sampled distribution in the tokenizer training data.

Taken together, our results suggest that extrinsic debiasing involves many interdependent challenges which cannot be inferred from an intrinsic outlook. We hope our work will promote more research on combining intrinsic debiasing methods to downstream tasks to produce extrinsically fairer MT models.

\section{Background}
\label{sec:background}



\label{debias_methods}

There is an abundance of debiasing methods in the field \cite{wang2021dynamically,schick2021self,shen2021contrastive,dev2019attenuating,dev2021oscar,kaneko2021debiasing,shao2022gold}. Most of them focus on intrinsic debiasing. We focus on three prominent methods, outlined below.
Importantly, all of these methods learn a transformation that can be applied to arbitrary vectors, once the model has finished training, and all were tested mostly intrinsically.

\paragraph{Intrinsic debiasing methods.}
We experiment with three methods: (1) \hd{} ~\citep{bolukbasi2016man} removes a gender subspace via a Principal Component Analysis (PCA) of predetermined word pairs which are considered as indicative of gender; 
(2) INLP~\citep{ravfogel2020null} 
learns the direction of the gender subspace rather than using a predefined list of words; and 
(3) LEACE~\citep{belrose2023leace} which prevents all linear classifiers from detecting a guarded concept.
A key difference between the methods is that \hd{} is non-linear and non-exhaustive, leaving stereotypical information after its' application \cite{gonen-goldberg-2019-lipstick}. In contrast, INLP and LEACE are linear and exhaustive; after applying INLP, stereotypical information can't be extracted with a specific linear classifier, and after applying LEACE, it can't be extracted with any linear classifiers.


\paragraph{The effect of debiasing on NMT.}
Most related to our work, \citet{escude-font-costa-jussa-2019-equalizing} explored the impact of debiasing methods on an English-to-Spanish MT task. However, they tested the MT models only on simple synthetic data, while here we focus on complex data reflecting real biases, and explore various design choices.

\section{Integrating Intrinsic Debiasing in MT}
\label{sec:methods}
We examine debiasing methods within the popular encoder-decoder approach to MT, as shown in Figure~\ref{fig:shwartz}.
Next, we describe the different research questions addressed in our setup. 

\paragraph{Which \emph{embedding} to debias?}
An encoder-decoder model has multiple embedding tables that can be intrinsically debiased: (1) the input matrix of the encoder; (2) the input matrix of the decoder; and (3) the output of the decoder, usually before the softmax layer.\footnote{In a complex system, such as the transformer encoder-decoder architecture, the representations after each transformer layer and within each layer can be debiased as well. We leave the investigation of such debiasing to future work.}
We employ different intrinsic debiasing techniques to each of these tables and evaluate their effect on downstream performance.
\paragraph{Which \emph{words} to debias?}
Tokenization poses a challenge for extrinsic debiasing as it may introduce discrepancies between the intrinsically debiased elements (complete words) and the MT input model (sub-word tokens) \citep{iluz2023exploring}. We experiment with three different configurations: (1) \emph{all-tokens}: debiases embeddings of all tokens in the model's vocabulary; 
(2) \emph{n-token-profession}: debiases all embeddings of words that appear in a predefined set of professions, even if they are split across multiple tokens, and (3) \emph{1-token-profession}: debiases only the embeddings of a predefined set of professions that align with the vocabulary of the debiasing technique, e.g., ``nurse'' is debiased only if it appears as a single token.
\paragraph{How does debiasing affect different \emph{languages}?}
We experiment with three target languages that encode morphological gender for nouns, representing different typological features: (1) Hebrew, a Semitic language with abjad script, (2) Russian, a Slavic language with a Cyrillic script, and (3) German, a Germanic language with Latin alphabet.

\section{Evaluation}
\label{sec:evaluation}


\subsection{Experimental Setup}
\label{sec:exp}

\begin{table}[tb!]
\centering
\small
 \begin{tabular}{llr} 
 \toprule
 Language & Dataset Name & Dataset Size \\ 
 \midrule
 Russian & newstest2019 & 1997 \\ 
 German & newstest2012 & 3003 \\
 Hebrew & TED dev & 1000 \\  
 \bottomrule
 \end{tabular}
 
 \caption{\label{tab:datasets}Datasets used for evaluating different target languages. The Dataset Size describes the number of sentences in the dataset.
 Russian and German datasets are described in \citet{Choshen2021TransitionBG}'s paper. The Hebrew dataset is based on the Opus TED talks dataset~\citep{reimers-2020-multilingual-sentence-bert}.}
\end{table}

\paragraph{MT model.}
We make use of 
\opus{}~\citep{TiedemannThottingal:EAMT2020},\footnote{\url{https://github.com/Helsinki-NLP/Opus-MT}} a transformer-based MT model built of 6 self-attention layers and 8 attention heads in the encoder and the decoder. The model was trained on Opus,\footnote{\url{https://opus.nlpl.eu}} an open-source webtext dataset, which uses SentencePiece tokenization~\citep{kudo-richardson-2018-sentencepiece}.
\paragraph{Metrics and datasets.}
For extrinsic debiasing measurement, we employ the automatic accuracy metric from \citet{stanovsky-etal-2019-evaluating}, assessing the percentage of instances where the target entity retains its original gender from the English sentence, using morphological markers in the target language.
We focus on the performance on the \emph{anti-stereotypical} set of 1584 sentences from WinoMT~\citep{stanovsky-etal-2019-evaluating}. These consist of anti-stereotypical gender role assignments, such as the female doctor in Example~\ref{ex1}.
In addition, we approximate the translation quality before and after debiasing using BLEU ~\citep{Papineni2002BleuAM} on several parallel corpora described in Table~\ref{tab:datasets}, and manually evaluate the translations to corroborate our findings.
Finally, all results are statistically significant with p-value $< 0.05$, see Appendix~\ref{sec:statistical_significance} for details.

\subsection{Results}

\begin{table}[]
\centering
\resizebox{\columnwidth}{!}{%
\begin{tabular}{lccc}
\toprule
Target Language & German & Hebrew & Russian \\ \midrule
no-debiasing & 57.7 & 45.6 & 41.0 \\ \midrule
n-token-profession & 60.9 & 48.3 & 41.0 \\
1-token-profession & \textbf{61.9} & \textbf{48.4} & \textbf{41.2} \\ \bottomrule
\end{tabular}%
}
\caption{Accuracy on different target languages when varying the tokens debiasing strategy. Presenting results for applying (1) the baseline (no-debiasing), (2) \textit{n-token-profession}, debiasing tokens corresponding to professions that are tokenized into one or more tokens, and (3) \textit{1-token-profession}, debiasing only professions that are tokenized into a single token. For brevity, each cell presents the the best performing choice of embedding table and debiasing method.
}
\label{tab:lang-eval}
\end{table}


\paragraph{Debiasing \emph{1-token-profession} professions outperforms other approaches.}
Table~\ref{tab:lang-eval} shows the gender translation accuracy when applying debiasing methods on different tokens. \footnote{Excluding results for debiasing all tokens, as it led to garbled translations where automatic debiasing measures are irrelevant.}
For the three tested languages, debiasing only professions that are tokenized into single tokens improved the gender prediction the most.
This hints that the sub-word tokens that compose a profession word do not hold the same gender information as the whole word.

\begin{table}[]
\centering
\resizebox{\columnwidth}{!}{%
\begin{tabular}{lcccc}
\toprule
\textbf{Embedding Table} & \textbf{Baseline} & \textbf{\hd{}} & \textbf{INLP} & \textbf{LEACE} \\ \midrule
Encoder Input            & 48.1 & \textbf{49.6} & 43.2 & 43.4          \\
Decoder Input            & 48.1 & 48.0 & 50.0 &\textbf{53.8}          \\
Decoder Output           & 48.1 & 48.0 & \textbf{50.7} & \textbf{53.8} \\ \bottomrule
\end{tabular}%
}
\caption{Opus MT's gender prediction accuracy with intrinsic debiasing methods applied on different embedding tables. Each cell is averaged across our target languages (\emph{de, he, ru}). Bold numbers represent best per debiasing method. The accuracy is measured by \citet{stanovsky-etal-2019-evaluating}'s method on their WinoMT dataset} 

\label{tab:embedding-table_opus_mt}
\end{table}

\paragraph{The optimal embedding table to debias depends on the debiasing method.} 
Table~\ref{tab:embedding-table_opus_mt} shows the improvement in gender prediction averaged across languages when applied on different embedding tables. 
\hd{} improves gender prediction only when debiasing the encoder's inputs, while INLP and LEACE improves gender prediction accuracy the most when applied to the decoder output. 
This may be explained by INLP's and LEACE's linearity, which therefore works best at the end of the decoder, after all nonlinear layers, while \hd{} employs a non-linear PCA component.\footnote{We tested debiasing all 8 combinations of the three embedding tables, but this did not change our findings.}


\paragraph{Results vary between languages.} 

Debiasing has a positive impact on the accuracy of gender translation in both German and Hebrew, with German improving by 3.7 points and Hebrew by 2.8 points. In contrast, Russian did not see as much improvement (Table~\ref{tab:lang-eval}).
The difference may be due to Russian's relatively rich morphology (e.g., it has 7 cases compared to 4 in German~\citep{wals}), resulting in much fewer single-token professions (59\% in Russian compared to 65\% in Hebrew, and 83\% in German). 



\begin{figure}[!t]
    \centering
    \includegraphics[width=1.0\columnwidth,center]{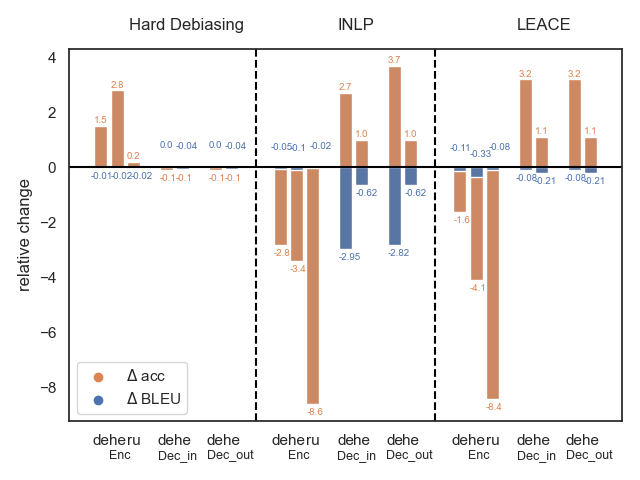}
    \caption{\label{fig:bleu_per_anti_accuracy}The relation between gender prediction accuracy difference (orange) and the BLEU difference (blue) between the original model (without any intervention) and the debiased model. The left part presents the results with \hd{},  INLP in the middle, and LEACE on the right. For each method, we present the results per each location (Encoder, Decoder-input, and Decoder-output), as well as each language).}
\end{figure}

\paragraph{LEACE and \hd{} do not significantly harm BLEU scores.}
Figure~\ref{fig:bleu_per_anti_accuracy} shows the relationship between the difference in the gender prediction and the difference in BLEU.
\hd{} and LEACE both have a small negative effect to the BLEU scores, while
in comparison, INLP presents a trade-off between the improvement in gender prediction and the translation quality according to BLEU scores. This shows that INLP removes information which is important for the translation model, while LEACE (which was proved to be the minimal transformation needed to remove gender information) and \hd{} indeed preserve more of the information. In terms of the gender prediction accuracy, the best setting of \hd{} is when applied to the encoder, while INLP and LEACE improve the gender prediction the most when applied to the decoder outputs. LEACE performs better than INLP when applied on the decoder as it was designed to prevent all linear classifiers from detecting the guarded concept, while INLP learns to obfuscate only one linear classifier.
\paragraph{Human evaluation shows that gender prediction is indeed improved with \hd{}.}
We manually annotate a portion of the translation to assess how well the automatic gender prediction metrics estimate real bias.
We annotate the configuration of \hd{} which changes the translations the most compared to other methods: applying \hd{} to the encoder's input with the \emph{1-token-profession} paradigm.
Out of the 1584 sentences in the dataset, 184 (11\%) changed after the debiasing. 32\% out of all the sentences that changed after the debiasing corrected the profession's gender prediction. These numbers are somewhat higher than what the automatic metrics suggest (26\% improvement on the same setup). See Appendix \ref{sec:human_evaluation} for additional details.






\section{Conclusions and Future Work}
\label{sec:conclusions}
We systematically explore different challenges and design choices when integrating intrinsic debiasing methods within complex machine translation systems. We find that it is better to debias only words representative of gender and correspond to single tokens,
that it is important to couple the debiasing method with the specific embedding table (e.g., encoder versus decoder), and that different target languages lead to vastly different results. Future work can evaluate additional debiasing methods on additional tasks, that may require other considerations when applying such methods.

\section*{Limitations}
Our work explores the integration of debiasing within a complex machine translation system. 
As such, the space of possible combinations to explore is very large, including the embedding table to debias, the choice of 
target languages, their corresponding test corpora, the debiasing method to explore and their hyperparameter settings, and more. 
We systematically explore a subset of these options, which may hinder the generalizability of our specific results, e.g., which tokenization scheme works best. We encourage future work to re-examine our findings in other settings and possibly refine or amend them, while our main takeaway is the broader set of considerations which should be taken into account when debiasing complex, real-world systems.
Additionally, to solve this task, machine translation systems need to also improve their coreference resolution abilities, which we did not examine here \cite{yehudai-etal-2023-evaluating}.
Additionally, our work focuses on gender bias, but certain debiasing techniques are broad and can be used for other protected attributes, thus we aspire that our work will pave the way for exploring other attributes in future works.


\bibliographystyle{acl_natbib}

\appendix




\section{Human Evaluation}
\label{sec:human_evaluation} 
The evaluation proposed by \citet{stanovsky-etal-2019-evaluating} is automatic and thus bound to introduce some noise. To mitigate this concern, we perform a human evaluation which complements the automatic metric and provides a more thorough analysis of the results.
The evaluation was applied to translations into Hebrew which were debiased according to the \emph{1-token-profession} words (see Section~\ref{sec:methods}), when specifically applied on encoder's input.
We only analyze the sentences that changed in any way compared to the original (non debiased) translations. 
There were 184 out of 1584 sentences which changed (11\%). 32\% of the changed sentences improved the gender prediction (i.e the gender of a profession was incorrect in the original translation and correct in the debiased translation), 3\% of the changed sentences were detracted (i.e the gender of a profession was correct in the original translation and incorrect in the debiased translation), and in the rest 65\% the change did not affect the gender of the profession. 
In 23\% out of the 65\% of unaffected sentences, the gender of the other profession in the sentence was changed, for example in the sentence ``The salesperson talked to the auditor because she was worried about the audit'' the gender of salesperson did not change in the debiased translations, but the gender of the auditor was changed from male to female form. Note that this is not a mistake since the gender of auditor is unknown in the source sentence. This shows that the debiasing method affects a larger amount of professions which are not counted in the improvement of the model.

\section{Human Annotations}
\label{sec:human_annotations} 
To define the gender direction in the target language for both debias methods, we needed the translations of the 10 representative gender word pairs for each language.
To get those pairs, we asked a native speaker of each of these languages to translate them into their language. In the case of a pair that is irrelevant to the target language (like Mary and John which are common male and female names in English but not in other languages), we asked them to adapt the pair to represent gender pairs in their language.
The set of professions that we debias was also translated into the target languages by three native speakers in each language. The professions annotations were taken from \citet{iluz2023exploring}.
The translations of the 10 pairs were collected for four languages, German, Hebrew, Russian, and Spanish. \footnote{link for the 10 pairs datasets will be released upon publication.}

\section{Statistical Significance}
\label{sec:statistical_significance} 
In order to determine the statistical significance of our findings, we employed McNemar's test, as recommended by \citet{dror-etal-2018-hitchhikers}.
McNemar’s test is designed for models with binary labels, therefore it is suitable to test the gender bias scores where each sentence is classified as correct if the gender is accurately identified in the translation and incorrect otherwise.
The null hypothesis for this test states that the marginal probability for each outcome is equal between the two algorithms being compared, indicating that the models are identical. In our case, the two models being compared are the original translation model and the debiased version.
When concatenating results per debias method, we get that the results of Hard Debias are significant with p-value of 3.01E-07, and the results of INLP are significant with p-value of 9.65E-06.
When comparing the results per embedding table to debias, we get that debiasing the encoder inputs is significant with p-value of 5.42E-10, debiasing the decoder inputs is significant with p-value of 0.016 and debiasing the decoder outputs is significant with p-value of 0.014.
finally when concatenating all the results, we get that comparing the outputs of a debiased model to a the original model, the results are significant with p-value of 0.01.

\end{document}